\documentclass{llncs}
\usepackage{amsmath,amssymb}
\usepackage{graphicx}
\usepackage{cite}
\usepackage{url}

\usepackage{booktabs}
\usepackage{multirow}
\usepackage{array}
\usepackage{amsmath}
\usepackage{tabularx}
\usepackage{csquotes}
\newcommand{\ra}[1]{\renewcommand{\arraystretch}{#1}}
\usepackage{color}

\begin{document}

\title{Domain Adaptation for Resume Classification Using Convolutional Neural Networks}
\author{Luiza Sayfullina, Eric Malmi, Yiping Liao and Alexander Jung}
\institute{Department of Computer Science, Aalto University, Finland}

\maketitle
\begin{abstract}
   We propose a novel method for classifying resume data of job applicants into 27 different job categories using convolutional neural networks. Since resume data is costly and hard to obtain due to its sensitive nature, we use domain adaptation. In particular, we train a classifier on a large number of freely available job description snippets and then use it to classify resume data. We empirically verify a reasonable classification performance of our approach despite having only a small amount of labeled resume data available. 
\end{abstract}

\section{Introduction}

The fast paced online job-market industry requires recruiters to screen through vast amounts of resume data in order to evaluate applicants fast and reliable. The design of accurate automatic classification systems typically requires the availability of labeled resume data which can be used to train the classifier.

Due to its sensitivity, resume data is difficult and costly to obtain. In contrast, data about job descriptions can be obtained much easier.
However, the two domains constituted by resume data and job description data are intrinsically related. Indeed, both data domains are related to the same job recommendation task, which is to match applicant to suitable job offers. Moreover, the resumes of applications have semantic similarities with job descriptions which belong to the same job category. For instance, they both can contain skills, education, duties as well as personal characteristics of the desired candidate.

So far, there are two main flavors along with their hybrids \cite{hong2013job}, of job recommendation systems. One class, referred to as content-based recommendation systems, is based mainly on the available job descriptions. A second class, referred to as collaborative filtering recommendation systems, is mainly based on the preferences of users who are interested in similar jobs. Content-Based recommendation system suggests to a user textually similar jobs to what he/she viewed or liked previously \cite{al2012survey}. 

It seems therefore reasonable to use transfer learning in order to implement a domain adaptation in order to leverage the information contained in vast amounts of labeled job description data in order to classify resume data. Since resume and job summaries belong to similar domains, we expect features extracted by a convolutional neural network for job classification to be highly relevant for resume summaries as well.


The theory of learning in different domains was theoretically approached in \cite{ben2007analysis,ben2010theory}. The authors provided generalization bound for domain adaptation using $\mathcal{H}$-divergence. It consists of two components and tries to find a trade-off between source-target similarity and source training error. Based on that assumption several researchers \cite{ganin2016domain,ajakan2014domain,ganin2014unsupervised} came up with the domain-adversarial approach, where high-level representations from neural network are optimized to minimize the loss on the source domain and maximize the loss on the domain classifier. \cite{long2015learning} proposed another approach based on convolutional neural network special architecture. First three layers are domain-invariant, next two layers are fined-tuned and fully connected layers aim to fit specific tasks, but regularized by multiple kernel variant of maximum mean discrepancy that enforces distributions to be similar. The proposed network is optimized for the image domain, being more specific.

Convolutional neural networks (CNN) have been successfully applied to not only image, but also text classification \cite{kim2015character,zhang2015character}, provided that enough training data is available. We propose a domain adaptation approach \cite{glorot2011domain,daume2006domain} where we train a CNN based classifier on 85,000 job description snippets which are labeled using 27 industrial job classes. After the classifier has been trained, we apply it to classify unlabeled resume data.


The paper is organized as follows. First, in Section~\ref{sec:dataset} we describe job, resume and children dream job datasets, used for classification. Then in Section~\ref{sec:industry}, we describe the \texttt{fastText} baseline model and the CNN for short text classification model. Experimental results are provided in Section~\ref{sec:exp}, where classification accuracies are reported along with t-SNE visualization built on latent CNN representations. Finally, we present conclusions in Section~\ref{sec:conclusions}.
 
\section{Datasets}\label{sec:dataset}
We study three different datasets: \textit{job descriptions}, which are used for training models, \textit{resume summaries}, which are our main target domain used for testing the models. \textit{Children's dream job} descriptions is rather a toy data lacking enough samples for fair evaluation, but these job descriptions significantly differ and thus are interesting to experiment with.


\subsection{Job Descriptions}

We collected 90,000 job description snippets using the Indeed Job Search API\footnote{\url{https://www.indeed.com/publisher}}, that enables access to short job summaries given a key word. As key words, we used 27 different industrial job categories listed in Table~\ref{T:dataset2}.

Here is an example of a job summary from the category \textit{Accountant}:			\blockquote{Entering journal entries, posting cash, and account reconciliations/supporting schedules. This position is responsible for supporting the daily operations and ...}.

Note that the snippets provided by Indeed are generated based on the full description of the job postings, thereby they encapsulate only the condensed information regarding the job. Furthermore, since the descriptions are unstructured text snippets, the contents provided by different companies for similar positions may be inconsistent. For example, some job snippets or summaries do not include informative sentences or keywords related to job titles or categories. However, since the descriptions are not limited by a predefined structure, they may provide richer and more detailed information about the jobs in varying industries.

\begin{table}[ht]
\ra{1.3}
\centering
\caption{27 industrial job categories from \texttt{https://www.indeed.com/find-jobs.jsp}.}
\label{T:dataset2}
\resizebox{\textwidth}{!}{%
\begin{tabular}{l l l}
\toprule
1. Accounting/Finance       &    10. Banking/Loans               &   19. Education/Training \\   
2. Healthcare               &    11. Human Resources             &   20. Legal               \\
3. Non-Profit/Volunteering  &    12. Restaurant/Food service     &   21. Telecommunications  \\  
4. Administrative           &    13. Construction/Facilities     &   22. Engineering/Architecture \\
5. Computer/Internet        &    14. Insurance                   &   23. Manufacturing/Mechanical \\
6. Pharmaceutical/Bio-tech  &    15. Retail                      &   24. Transportation/Logistics  \\
7. Arts/Entertainment/Publishing & 16. Customer Service          &   25.Government/Military \\ 
8. Hospitality/Travel       & 17. Law Enforcement/Security       &   26. Marketing/Advertising/PR \\           
9. Real Estate              & 18. Sales                          &   27. Upper Management/Consulting \\    
\bottomrule
\end{tabular}
}
\end{table}

\subsection{Resume Summaries}

\begin{figure}
    \centering
\includegraphics[width=0.8\linewidth]{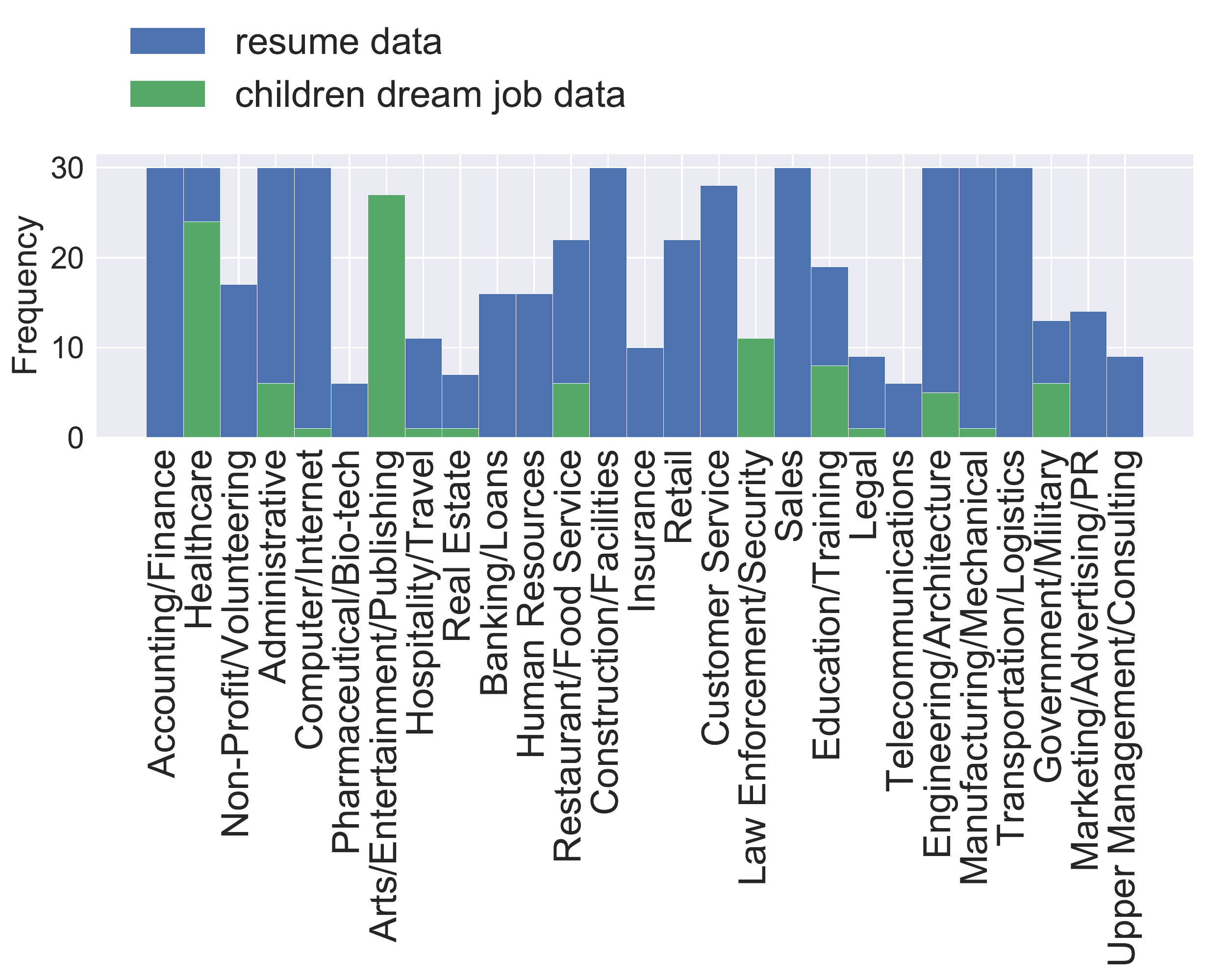}
\caption{The comparison of class distribution in job description and resume datasets.}
    \label{fig:category_hist}
\end{figure}

We collected 523 anonymous resume data samples, each sample labeled with one of the 27 categories based on the type of a job the candidate is looking for. The distribution of the categories is shown in Figure~\ref{fig:category_hist}. 

Here is an example of a resume self-description summary:
\blockquote{``Experienced analyst with an excellent academic profile and having several years of invaluable experience in domestic and international consultancy and management. Highly focused with a comprehensive knowledge and understanding of project management, technical issues and financial practices. Good at meeting the deadlines. Consider myself to be sociable person and good team worker.''}

\subsection{Children's Dream Jobs}

Children, unlike grown-ups, can express their dream jobs more emotionally, without being attached to skills, but rather following their interests. So in addition to resumes, we decided to use children dream job descriptions that were categorized manually into the same 27 job categories. The data set contains 98 children's short essays on their dream job parsed from \footnote{\url{http://www.valleymorningstar.com/sie/what_do_you_think/article_692e1ac9-bae5-5705-8005-c22dac04ebf6.html}}.  Below is an example essay:
\blockquote{``As far as I can remember I have always wanted to become a medical doctor. More specifically, a cardiologist. I love the thought of saving a person's life. The road to becoming a doctor is a long process, but worth it in the end. Having the feeling of accomplishment and knowing that I have made an impact on a family's life, would be the greatest satisfaction for me.''}

\subsection{Comparison of Job Descriptions and Resumes}

Since our aim is to leverage easily available job description data to train a model for classifying resume summary snippets, it is important to understand how these two domains differ from each other. In order to compare the two, we study word frequencies to see whether certain terms are over-represented in one domain compared to the other.

\begin{figure}
    \centering
    \includegraphics[width=0.7\linewidth]{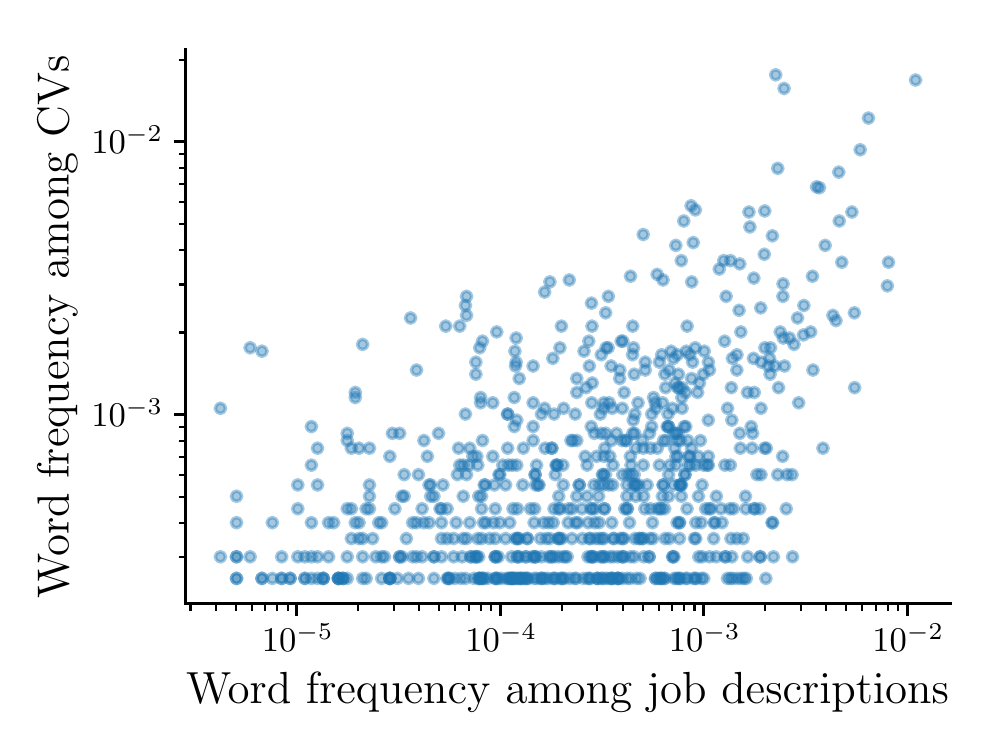}
    \caption{Normalized frequencies of all the words appearing at least five times in both datasets. Each word corresponds to a dot whose $x$ and $y$ coordinates denote the frequencies among job descriptions and CVs (resumes), respectively.}
    \label{fig:word_freqs}
\end{figure}

Figure~\ref{fig:word_freqs} shows the normalized frequencies of all words appearing at least five times in both datasets. The two frequencies are correlated ($\rho = 0.59$), but we can see, that for some words, the frequencies differ considerably. In Table~\ref{tab:word_freqs}, we list the words for which the relative difference is the largest.\footnote{The relative difference is measured by dividing the two normalized frequencies. For low frequencies, this measure will be noisy but we ignore this since the purpose of the experiment is to merely gain an overview of the differences between the datasets.} The results show that in resumes people are much more likely to use adjectives describing themselves, such as \textit{adaptable} and \textit{polite}, whereas job descriptions mention more often roles, such as \textit{director} and \textit{coordinator}.

\begin{table}[ht]
\centering
\caption{Words which normalized frequency differs the most between job descriptions ($f_{\textmd{Job}}$) and resume summaries ($f_{\textmd{CV}}$). Difference is measured by dividing the frequencies.}
\label{tab:word_freqs}
\begin{tabular}{l l l l l}
\toprule
   & $f_{\textmd{CV}} / f_{\textmd{Job}}$ & Word & $f_{\textmd{Job}} / f_{\textmd{CV}}$ & Word \\ \midrule
 1. & 284.9 & uk                   & 15.3 & program \\
 2. & 242.2 & gained               & 10.4 & assist \\
 3. & 239.3 & adaptable            & 9.0 & director \\
 4. & 95.0 & polite               & 8.4 & medical \\
 5. & 82.1 & keen                 & 6.7 & provides \\
 6. & 76.0 & bsc                  & 6.7 & coordinator \\
 7. & 73.3 & trustworthy          & 6.6 & accounting \\
 8. & 73.3 & ambition             & 6.5 & executive \\
 9. & 63.3 & licence              & 6.2 & representative \\
10. & 59.6 & confident            & 5.9 & assistant \\
11. & 59.5 & adapt                & 5.8 & report \\
12. & 57.0 & versatile            & 5.7 & food \\
13. & 57.0 & consultancy          & 5.7 & perform \\
14. & 52.9 & approachable         & 5.7 & equipment \\
15. & 48.8 & punctuality          & 5.6 & related \\
\bottomrule
\end{tabular}
\end{table}

\section{Industrial Category Classification Methods}\label{sec:industry}

The objective of industrial category classification is to classify user profiles, represented as text snippets, into 27 industrial categories shown in Table~\ref{T:dataset2}. We apply a CNN based methods to this task because they have shown state-of-art performance in text classification \cite{cnn_sentence}.
As a baseline method, we employ the \texttt{fastText} classifier~\cite{joulin2016bag} which is presented next.

\subsection{Fast text Classifier}
\label{sec:fastText}

The \texttt{fastText} method has been proposed recently by Joulin et al.~\cite{joulin2016bag} to efficiently classify text data. The method is based on learning word embeddings, averages them, and feeds the resulting vector into a linear classifier. The method also supports learning word embeddings for n-grams which allows capturing word order information.

Supported by a few algorithmic and implementation improvements, \texttt{fastText} is able to train and test extremely fast without access to GPUs. We have chosen fastText, since it was shown by Joulin et al. \cite{joulin2016bag} to be a competitive baseline for deep learning models, outperforming models like CNN, char-CNN and slightly (~1\%) underperforming LSTN-GRNN models. 


\subsection{Convolutional Neural Networks for Sentence Classification}
\label{s_cnn}

Word2vec model \cite{mikolov2013distributed} is a widely used method for learning vector representations of words, so that semantically similar words are close to each other in the vector space.  Based on the word vectors, contextual information can be extracted to learn the semantic similarity between words and sentences. 

Convolutional neural networks trained on the top of pre-trained word2vec representations proposed by \cite{cnn_sentence} showed state-of-the-art performance on several datasets, including sentiment analysis.
In this model, words in the sentences are embedded word2vec representation of the same length. Then vectors of words are concatenated by rows thus forming a matrix, to which CNN is applied. 

Let us introduce some notations. First,  $\mathbf{x_i} \in \mathbb{R}^{k}$ is the $i$-th word embedded into a vector of $k$ dimensions. A sentence which consists of $n$ words is represented as $\mathbf{x_{1:n}} = \mathbf{x_1}\bigoplus 
\mathbf{x_2} \cdots \bigoplus \mathbf{x_n} $, where $\bigoplus$ is the concatenation operator. The 
convolutional filter with window size of $h$ words is denoted as $\mathbf{w} \in \mathbb{R}^{hk}$.
Then the new feature map is generated by:
\begin{equation}
 c_i = f(\mathbf{w} \cdot x_{i:i+h-1}+b),
\end{equation} 
where $b \in \mathbb{R}$ is a bias term and $f$ is a hyperbolic tangent. Following the convolution operation, 
the max-pooling operation is applied to capture the most important feature and the output is forwarded to a 
fully connected soft-max layer whose output is a probability distribution over classes. The regularization of the network is done by applying dropout to prevent co-adaptation and re-scaling weights to prevent large (and possibly noisy) gradient updates during training. 
At the testing phase, the learned weight vectors are scaled by $\mathbf{w} \leftarrow p\mathbf{w}$. Additionally, an
L2-norm constraint is applied to rescale $\mathbf{w}$ to have $||\mathbf{w}||_{2} = s$ when $||\mathbf{w}||_{2} > s$ after gradient decent step.

\section{Experimental results}\label{sec:exp}

We trained our models by fixing training (80,000 samples) and validation data (5,000 samples) consisting of job summaries and used all available samples from resume and children's dream job data for testing. We also used 5,000 job summary samples for testing a classifier on purely job data. All selected data samples were trimmed to 100 words.

Our  CNN  model  was  based  on  the  implementation  by  Kim \footnote{\url{https://github.com/yoonkim/CNN_sentence}}.  In  order  to avoid strong overfitting, we increased the L2-norm constant up to 10 and set the width of CNN filters to be [2,3,4] instead of [2,3,4,5]. We tried the filters of size [50, 100, 200] per each type and have chosen 50 based on validation set from job description data.  The  dropout  rate  was  set  to  0.5  and  we  found  it  useful  for  model regularization.  A  non-static  setting  of  CNN  was  chosen,  where  Google-News pretrained word vectors are fine-tuned while training.


For the \texttt{fastText} model, we optimized the lengths of the n-grams and the learning rate hyperparameters using the validation data, obtaining the values 4 and 0.25, respectively. These values were kept fixed for all three test datasets.

The overall prediction accuracies are shown in Table~\ref{tab:results}. When moving from the source domain to the target domain, the accuracy drops from 74.88\% to 40.15\%. CNN outperforms \texttt{fastText} for each dataset and particularly for resume and dream job data, which shows that the CNN model generalizes better to new domains. 

\begin{table}[ht]
\caption{Job category prediction accuracies (\%) for the fastText method and CNN for short text classification. \label{tab:results}}
\centering
\begin{tabularx}{\textwidth}{XXX}
\toprule
\textbf{Dataset} & \textbf{fastText} & \textbf{CNN}  \\ \midrule
Job description & 71.99 &  \textbf{74.88} \\ 
Resume & 33.40 & \textbf{40.15} \\ 
Children's dream job & 28.5 & \textbf{51.02} \\ 
\bottomrule 
\end{tabularx}
\end{table} 

\begin{figure}[htp!]
    \centering
    \includegraphics[width=1\linewidth]{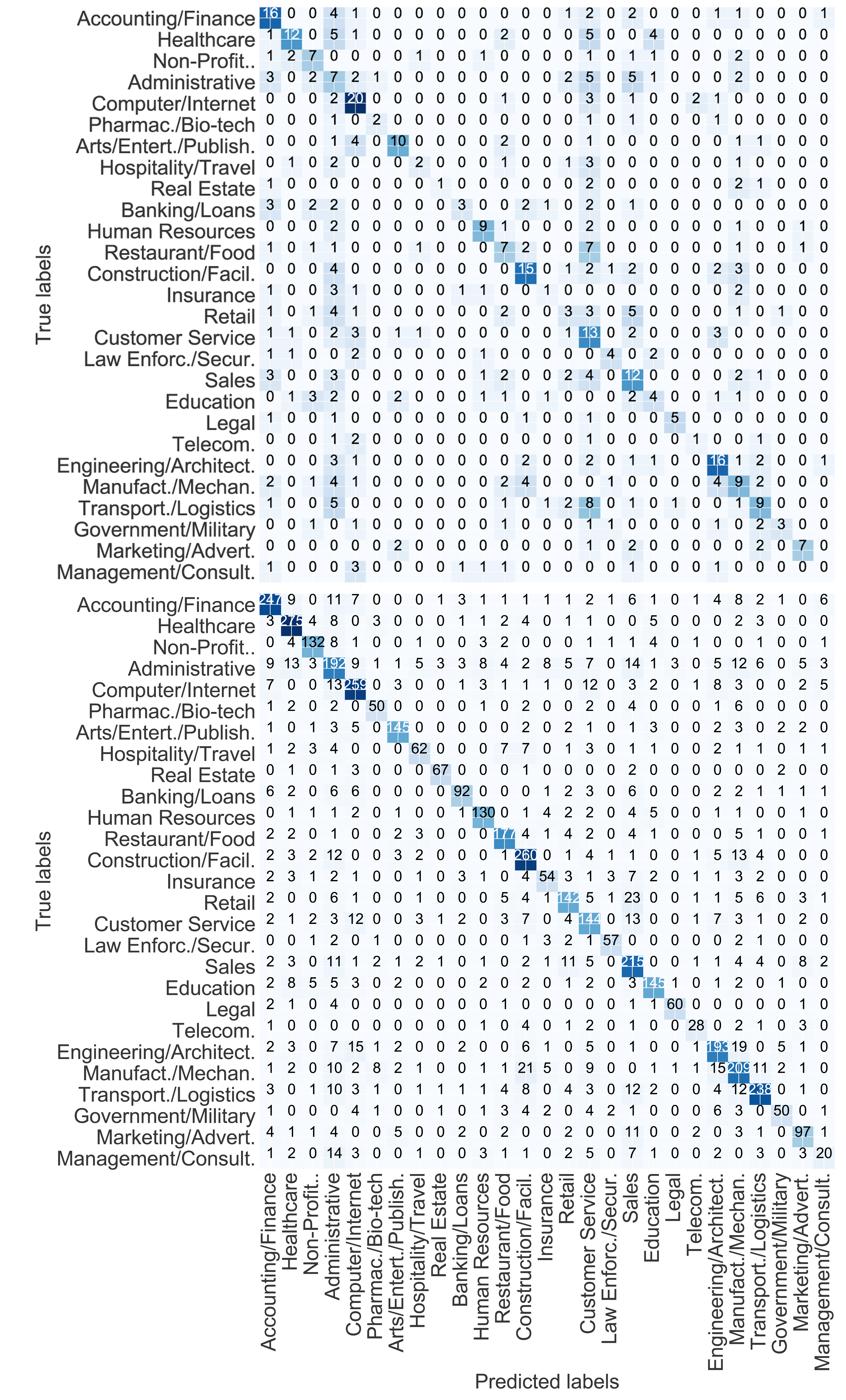}
    \caption{Confusion matrix of resume (top) and job (bottom) classification results. In both datasets Management, Administrative, Customer Service, Retail and Manufacturing categories have a low recall. We assume that this happens due to the semantic closeness of these categories, since even a human can not always correctly make a clear distinction between them.}
    \label{fig:conf_matrix}
\end{figure}

The confusion matrices for job description and resume summary classification are shown in Figure~\ref{fig:conf_matrix}.

%

\begin{figure}[!htp]
    \centering
    \includegraphics[width=1\linewidth]{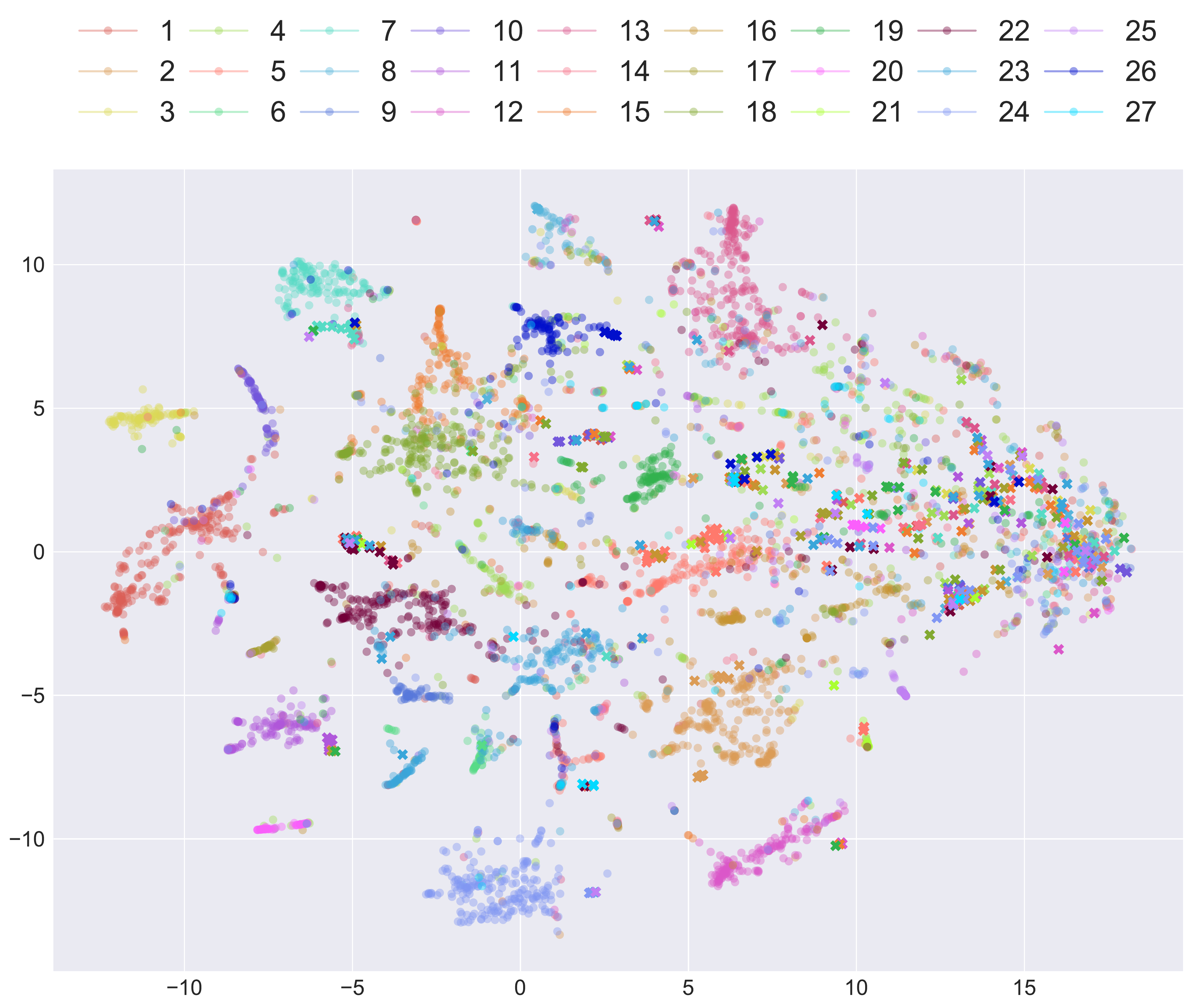}
    \caption{t-SNE visualization using the CNN first layer outputs on job and resume data. We used all job (5,000) and resume (523) test data for fitting t-SNE. By visualizing vectors on 2D space, we check how useful are the representations learned by CNN to distinguish between the classes in familiar and new domains. There are 5000 test samples from job data, marked with circles, and 523 samples from resume data, marked with crosses, in total. We can observe the presence of category clusters formed by job samples, however they are not perfectly separable. Since the resume data has differences in underlying distribution, some resume clusters are neighbours with corresponding job clusters, e.g from Non-Profit, Computer/Internet, Arts, Retail and Engineering categories. In fact, resume classes from these classes form neighbouring clusters or intersect with corresponding job clusters.}
    \label{fig:tsne}
\end{figure}

From Figure~\ref{fig:conf_matrix} we can see that the hardest categories to classify in job description dataset are Management, Administrative, Sales, Customer Service and Manufacturing. Probably, it happens due to semantic closeness of some job categories, like Management and Administrative, Sales and Retail, since even humans can have trouble clearly distinguishing between them. Manufacturing category samples were classified with Construction, Engineering and Transportation/Logistics labels. The highest recall belongs to Legal, Real Estate, Arts, Law and Non-Profit categories. 

 Resume dataset has a small number of samples per class, so we can not make general conclusions from confusion matrix showed in Figure~\ref{fig:conf_matrix}. Still, the results on our resume data have common trends with job data. For example, similarly to job confusion matrix, Management, Administrative, Customer Service, Retail and Manufacturing categories have a low recall. Legal, Government, Arts, Healthcare and Pharmacy show the highest recall. Management category, consisting of 9 samples, was not detected at all, probably since this position can be quite general and related to various job fields.
 
 One of the ways to achieve better generalization is building latent representations. In our case, the concatenated outputs $c_i$ of the first layer of the CNN model form latent space representations. Therefore, we visualized those outputs both for job and resume test data using t-SNE \cite{maaten2008visualizing} projection and show the results in Figure~\ref{fig:tsne}.
 
 One can observe the presence of category clusters formed by job samples, although some of them are not perfectly separable. However, for 27 classes this is relatively good separation. If resume samples, represented by crosses, were semantically close to job descriptions, they could belong to the same job clusters. However, since the resume data has differences in the underlying distribution, some of its clusters are at least neighbours with corresponding job clusters, e.g, for Non-Profit, Computer/Internet, Arts, Retail and Engineering categories. We can not make any general conclusions about resume clusters due to the lack of data, but we can clearly find clusters for some categories, that are sometimes distant from corresponding job clusters. This suggests that the learned CNN representations are useful for resume classification as well, since clusters can be found using them.
 

\section{Conclusion}\label{sec:conclusions}

We have devised a resume classification method which is able to exploit the information contained in vast amounts labeled job description data in order to achieve higher accuracy. Since resumes are more sensitive data and difficult to obtain, compared to job summaries, we trained the proposed model only on job summaries and tested its performance on resume data with the same job category labels. A convolutional neural network for short text classification using word embeddings was trained and validated on 85,000 short job summaries mined from Indeed. Then this network was used to classify a set of 523 candidate resumes and compared with a simple but effective \texttt{fastText} model. Our method achieved 74.88\% accuracy on job classification task and 40.15 \% on resume classification, thereby outperforming the existing \texttt{fastText} model by more than 6\% on resume classification task and 3\% on the job description task. Moreover, we applied our method to a small imbalanced dataset consisting of 98 children dream job descriptions. In this task CNN outperformed \texttt{fastText} by 22\%. 

Given the fact that no labels were used from resume data for training or validation, we consider CNN for short classification to be useful in a domain adaptation scenario. An interesting direction for future work would be to study whether the results can be improved by leveraging a small number of labeled resume samples to fine-tune the CNN model.


\bibliographystyle{splncs03.bst}
\bibliography{bibl}



\end{document}